\documentclass{article}
\usepackage{iclr2021_workshop,times}

\usepackage{amsmath,amsfonts,bm}

\def\eqref#1{equation~\ref{#1}}

\def\1{\bm{1}}

\DeclareMathAlphabet{\mathsfit}{\encodingdefault}{\sfdefault}{m}{sl}
\SetMathAlphabet{\mathsfit}{bold}{\encodingdefault}{\sfdefault}{bx}{n}

\usepackage{graphicx}
\usepackage{hyperref}
\usepackage{url}
\usepackage{adjustbox}

\newcommand{\relu}{\mathrm{RELU}}

\title{Simplicial Complex Representation Learning }

\author{Mustafa Hajij \\
Department of Mathematics and Computer Science\\
Santa Clara University\\
\texttt{mhajij@scu.edu} \\
\And
Ghada Zamzmi \\
Department of Computer Science \\
University of South Florida \\
\texttt{ghadh@mail.usf.edu} \\
\And
Theodore Papamarkou \\
Department of Mathematics \\
The University of Manchester\\
\texttt{theodore.papamarkou@manchester.ac.uk} \\
\And
Vasileios Maroulas \\
Department of Mathematics \\
University of Tennessee\\
\texttt{vmaroula@utk.edu} \\
\And
Xuanting Cai \\
Facebook Inc \\
\texttt{caixuanting@fb.com} \\
}

\iclrfinalcopy 
\begin{document}

\maketitle

\begin{abstract}
Simplicial complexes form an important class of topological spaces that are frequently used in many application areas such as computer-aided design, computer graphics, and simulation. Representation learning on graphs, which are just 1-d simplicial complexes, has witnessed a great attention in recent years. However, there has not been enough effort to extend representation learning to higher dimensional simplicial objects due to the additional complexity these objects hold, especially when it comes to entire-simplicial complex representation learning. In this work, we propose a method for simplicial complex-level representation learning that embeds a simplicial complex to a universal embedding space in a way that complex-to-complex proximity is preserved. Our method uses our novel geometric message passing schemes to learn an entire simplicial complex representation in an end-to-end fashion. We demonstrate the proposed model on a publicly available mesh dataset.  To the best of our knowledge, this work presents the first method for learning simplicial complex-level representations.


\end{abstract}


\section{Introduction}
Object representation learning aims to learn a mapping that embeds the elementary components of this object into some Euclidean space while preserving the object's structural information. Recently, such methods have gained a great momentum especially with graph representation learning. The latter has attracted considerable popularity over the past few years with success in both node-level representation learning \cite{cui2018survey} and entire graph learning \cite{narayanan2017graph2vec,tsitsulin2018netlsd}. The applications of such representation on graphs are diverse as they can be used for almost any downstream machine learning task on domains such as graph classification \cite{hamilton2017inductive} or similarity \cite{heimann2018regal}.


Despite the success of graph representation learning in the past few years, there is a gamut of applications, e.g. in neuroscience \cite{TangGiusti2019,Nasrin2019,Maroulas2019,Oballe2021} where learning in cliques of nodes and how the information passes among these cliques are urgently required, there has not been enough efforts to extend representation learning to simplicial complexes. In general, topological considerations of the underlying data and their relationships encapsulate rich information which one may harness, e.g. see \cite{Love2021,Maroulas2021}.  The higher dimensional simplicial complexes often hold additional structure over graphs that might be critical in modeling and must be incorporated to learn the correct representation.
For instance, when a simplicial complex is a triangulated manifold,
then the aim is to design an algorithm that uncovers the manifold in the learnt representation.
Motivated by the success of graph representation learning, we propose a method for learning simplicial complex representation. Our method utilizes a complex autoencoder proposed in \cite{hajijcell} and learns an entire simplicial complex representation extracted from simplices embeddings vectors induced by the simplicial complex autoencoder. Our learning function maps every simplicial complex to a universal embedding space in a way that complex-to-complex proximity is preserved. Learning simplicial complex-level representation is essential to perform downstream machine learning tasks on these objects such as simplicial complex classification and similarity ranking. See for instance \cite{hajij2018visual,fey2019fast,ying2018hierarchical,narayanan2017graph2vec} for related studies on graphs.



%

The literature of entire graph representation learning is rich and many methods have been proposed including Laplacian-based methods \cite{de2018simple,tsitsulin2018netlsd}, implicit factorization techniques \cite{chen2019gl2vec,narayanan2017graph2vec}, GNN-based methods
\cite{bai2019unsupervised}, and pooling based methods \cite{ying2018hierarchical}. We refer the reader to \cite{cui2018survey} for a recent survey on network embedding. In addition, simplicial complex representation learning, inspired by the success of node2vec \cite{angles2008survey} and Word2Vec \cite{mikolov2013efficient}, started to get attention recently. For example, the works in \cite{billings2019simplex2vec,schaub2020random} define simplicies emebddings based random walks on simplicial complexes. This was generalized to $k$-simplex embeddings in \cite{hacker2020k}. A general cell complex autoencoder scheme that describes these random walk-based representations as special cases was suggested in \cite{hajijcell}.

Although there are several methods for learning simplex-level representation \cite{billings2019simplex2vec,schaub2020random,hajijcell,hacker2020k}, the work herein is the first to propose a learning representation of the entire simplicial complex. Further, we show our simplicial complex autoencoder (SCA) on a publicly available mesh dataset and demonstrate promising preliminary results. The rest of the paper is organized as follows. Notation and necessary definitions on simplicial complexes are given in Section \ref{background}. Section \ref{cxn} is devoted to reviewing cell complex neural networks. Our proposed simplicial complex autoencoder (SCA) is given in Section \ref{autoencoders}. Finally, Section \ref{test} shows the results.

\section{Simplicial Complex Neighborhood Matrices}
\label{background}

This section provides the necessary notations to define neighborhood matrices between simplices in a simplicial complex, and hence, we assume the reader has familiarity with basic definitions of simplicial complexes \cite{hatcher2005algebraic}. Let $X$ be a simplicial complex and $n$ be the dimension of $X$. Recall that the dimension of $X$ is the dimension of the highest simplex in $X$. For any $0\leq k \leq  n$, we denote the set of all $k$-simplices in $X$ by $X^k$. If $X$ is a simplicial complex of dimension $n$, then for every $ 0 < m \leq n $ we denote the set of simplicies in $X$ with dimension less than $m$ by $X^{<m}$. The set $X^{>m}$ is defined similarly. In this work, we assume that the complex $X$ is unoriented. However, the following notion of neighbored on simplicial complexes can be easily extended to oriented simplicial complexes; see for instance \cite{hajijcell,glaze2021principled} for various considerations on oriented simplical complexes in the context of neural network computations.

Adjacency relations can be defined on simplicial complexes in a similar fashion as they are defined on graphs. Specifically, let $X$ be a simplicial complex and let $c^n$ denotes a $n$-simplex in $X$, and $facets(c^n)$ denotes the set of all $(n-1)$-simplicies $X$ incident to $c^n$. Two $n$-simplices $a^n$ and $b^n$ are said to be \textit{adjacent} if there exists an $(n+1)$-simplex $c^{n+1}$ such that $a^n,b^n \in facets(c^{n+1})$. The set of all simplices adjacent to a simplex $a$ in $X$ is denoted by $\mathcal{N}_{adj}(a)$. 
   Dually, $a^n$ and $b^n$ are \textit{coadjacent} in $X$ if there exists an $(n-1)$-simplex $c^{n-1}$ with  $a^n,b^n \in cofacets(c^{n-1})$. The set of all cells adjacent to a simplices $a$ in $X$ is denoted by $\mathcal{N}_{adj}(a)$ while the set of all simplices co-adjacent to a simplex $a$ in $X$ is denoted by $\mathcal{N}_{co}(a)$. If $a^n,b^n $ are $n$-simplices in $X$, then we define the set $\mathcal{CO}[a^{n}, b^{n} ]$ to be the intersection of $cofacets(a^n) \cap cofacets(b^n)$.  Similarly, the set $\mathcal{C}[a^{n}, b^{n} ]$ is defined to be the intersection of $facets(a^n) \cap facets(b^n)$. Observe that these notions generalize the analogous notions of adjacency/co-adjacency matrices on graphs.  Precisely, let $X$ be a simplicial complex of dimension $n$, $N$ be the total number of simplices $X$, and define $\hat{N}:=N-|X^{n}|$. Let $c_1,\cdots , c_{\hat{N}}$ denotes all the simplices in $X^{<n}$. The \textit{ adjacency matrix} of $X$, denoted by $A_{adj}$, is a matrix of dimension $\hat{N} \times \hat{N}$ and defined by setting $A_{adj}(i,j)=|\mathcal{CO}[c_i, c_j]|$ if the simplex $c_i$ is adjacent to $c_j$ and zero otherwise. We denote the adjacency matrix between $k$-simplices in $X$ by $A^k_{adj}$, where $ 0 \leq  k< n$. The co-adjacency matrices $A_{co}$, $A^k_{co}$ are defined dually by  storing $|\mathcal{C}[c_i, c_j]|$ where the simplices $c_i$ and $ c_j$ are co-adjacent.
  




\section{Geometric Message Passing Schemes (GMPS) }
\label{cxn}

\begin{figure*}[htb]
  \centering
   {\includegraphics[scale=0.024]{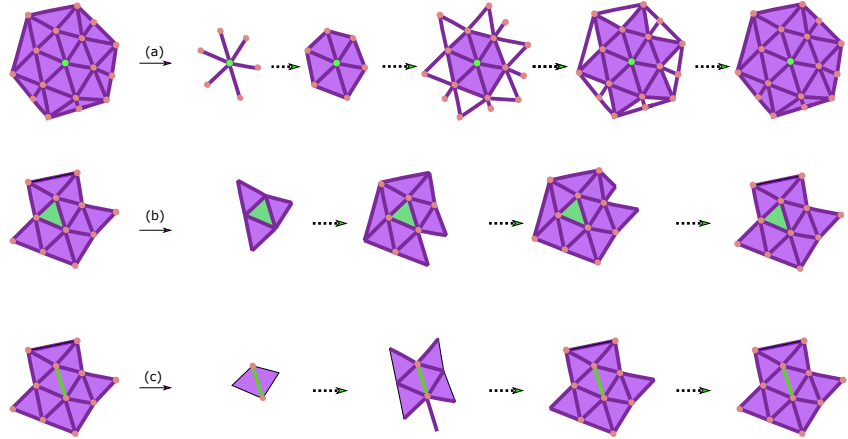}
    \caption{Illustration of the flow of information with various geometric message passing schemes (GMPS). (a) Flow of information with adjacency message passing schemes (AMPS). (b) Flow of information with co-adjacency message passing scheme (CMPS). (c) Flow of information with homology and cohomology message passing scheme (HCMPS). In all examples, the computations flow is given from the green simplex. }
  \label{111}}
\end{figure*}

This section briefly reviews the basic definitions and notations of cell complex networks (CXN) introduced in \cite{hajijcell} as it is applicable in our context on simplicial complexes. Specifically, every simplicial complex $X$ is a cell complex where the $k$-simplexes $X$ in that complex are precisely the $k$-cells. In what follows, we will use the terms ``cell'' and ``simplex'' interchangeably to refer to simplices in a given complex $X$. In \cite{hajijcell} three general message passing schemes on cell complexes were suggested. These schemes are trivially applicable in our case here on simplicial complexes. In order to implement a specific neural network of a complex, here a simplicial complex autoencoder, one must select a particular message passing scheme or a combination of them. This consequently affects the final representation of the simplicial complex. In Section \ref{cxn} we mentioned one possible geometric message passing scheme which is AMPS.  In this section we review briefly these \textit{geometric message passing schemes} (GMPS) on a general cell complex net (See Figure \ref{111} for an illustration of the flow of data computations with these schemes). Note that the simplicial complex autoencoder given in Section \ref{autoencoders} employs AMPS. Other simplicial complex autoencoders can be defined similarly using the message passing schemes that we shall present.

\subsection{Adjacency Message Passing Scheme (AMPS)}

The input for an AMPS-CXN is specified by cell embeddings $H^{(0)}_m \in \mathbb{R}^{ |X^m| \times d_0}$ that define the initial cell features on every $m$-cell in $X$. Here, $d_0$ is the dimension of the input feature embedding dimension of the cells. Given the desired depth $L>0$ of the CXN net one wants to define on the complex $X$, the \textit{adjacency message passing scheme} (AMPS) on $X$ consists of $L\times n $ cell embeddings and it is defined as
\begin{equation}
\label{CCXN}
    H^{(k)}_m:=M \left( A_{adj}, \mathbf{\partial}, H^{(k-1)}_m,H^{(k-1)}_{m+1} ;\theta^{(k)}_m \right) ,
\end{equation}
where $0 \leq m \leq n-1$, $1 \leq k \leq L$,  $H^{(k)}_m \in \mathbb{R}^{|X^m| \times d_k}  $ are the cell embeddings computed after $k$ steps of applying Equation \eqref{CCXN}, $\mathbf{\partial}$ are the boundary operators needed for the computations, and $\theta^{(k)}_m$ is a trainable weight vector at the layer $k$, $M$ is the message propagation function that depends on the weights  $\theta^{(k)}_m $, the cell embeddings $H^{(k)}_m$ and the adjacency matrix of $X$.  The propagation function $M$ can be implemented in many ways. Observe that the matrix $A_{adj}$ can be replaced by a version of the k-Hodge Laplacian matrix as well \cite{roddenberry2021signal}. For instance, in \cite{hajijcell} a generalization for graph convolutional neural networks \cite{kipf2016semi} to convolutional cell complex networks was provided.  Finally, see also \cite{ebli2020simplicial} for related implementations on simplicial complexes. 

\begin{figure*}[ht]
  \centering
   {\includegraphics[scale=0.04]{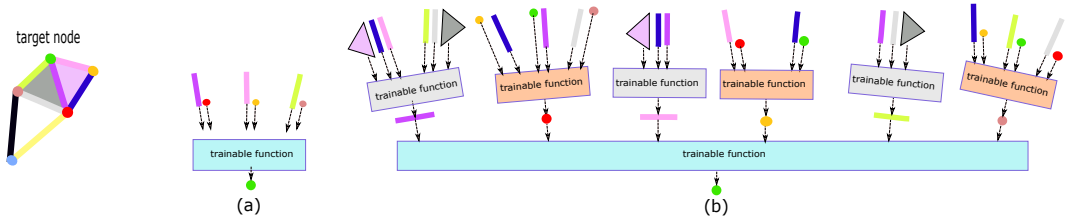}
    \caption{Two layers cell complex neural network (CXN) \cite{hajijcell} illustrated on a simplicial complex. The computations are  illustrated with respect to the green target vertex.  }
  \label{DMF1}}
\end{figure*}

 Note that the information flow using Equation \eqref{CCXN} on the complex from the lower dimensional cells to the higher ones. Further, note that the message passing scheme given by Equation \eqref{CCXN} does not update the feature vectors associated with the final $n-$ cells on the complex. If such a property is desirable, then Equation \eqref{CCXN} must be adjusted using co-adjacency information of the simplicial complex. See \cite{hajijcell} for details. Figure \ref{DMF1} demonstrates how the cell embeddings are updated.



\subsection{Co-adjacency Message Passing Scheme (CMPS)}
\begin{figure}[ht]
  \centering
   {\includegraphics[scale=0.0465]{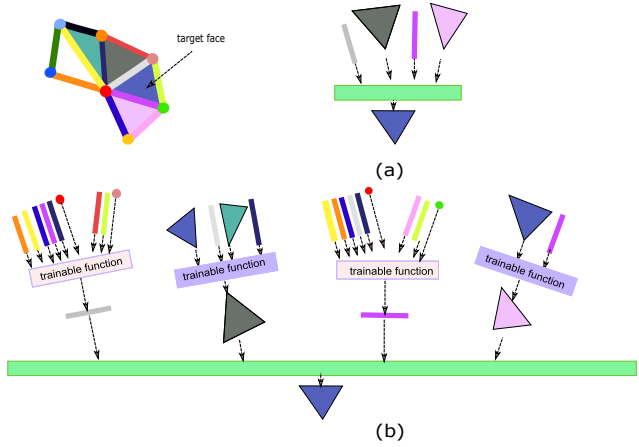}
    \caption{Coadjacency message passing scheme. The above illustrates the computation of a two layer cell complex network with respect the specified target face.  }
  \label{DMF2}}
\end{figure}

Co-adjacency message passing scheme is very similar to the AMPS we  mentioned in Section \ref{cxn}. The only difference is that it utilizes the co-adjacency relations of a given face instead of the adjacency matrix. Specifically, let $H^{(0)}_m \in \mathbb{R}^{ |X^m| \times d_0}$ be the initial cell feature vector on every $m$-cell in $X$. Let $L>0$ be the desired depth of the CXN one wants to define on a complex $X$, the \textit{Coadjacency Message Passing Scheme} (CMPS) on $X$ consists of $L\times n $ embeddings and it is defined as
\begin{equation}
\label{xxx}
    H^{(k)}_{n-m}:=M\left( A_{co},\mathbf{\partial}^T, H^{(k-1)}_{n-m},H^{(k-1)}_{n-m-1} ;\theta^{(k)}_{n-m}\right) , 
\end{equation}
where $0 \leq m \leq n-1$, $1 \leq k \leq L$,  $H^{(k)}_{n-m} \in \mathbb{R}^{|X^{n-m}| \times d_k}  $ are the embeddings computed after $k$ steps of applying Equation \eqref{xxx},  $\mathbf{\partial}^T$ are the coboundary operators needed for the computations, and $\theta^{(k)}_{n-m}$ is a trainable weight vector at the layer $k$, $M$ is the message propagation function that depends on the weights  $\theta^{(k)}_{n-m} $, the cell embeddings $H^{(k)}_{n-m}$ and the adjacency matrix of $X$.  Observe also that the matrix $A_{adj}$ can be replaced by a version of the k-Hodge Laplacian matrix as well \cite{roddenberry2021signal}. See Figure \ref{DMF2} for an illustration of such message passing scheme on a simplicial complex.

Note that with CMPS, the flow of information goes from higher cells to lower ones. This explains the strange index choice in Equation \eqref{xxx}. Moreover, note that the feature vectors associated with the zero-cells are never updated.

\subsection{Homology and Cohomology Message Passing Scheme (HCMPS)}
\begin{figure}[ht]
  \centering
   {\includegraphics[scale=0.05]{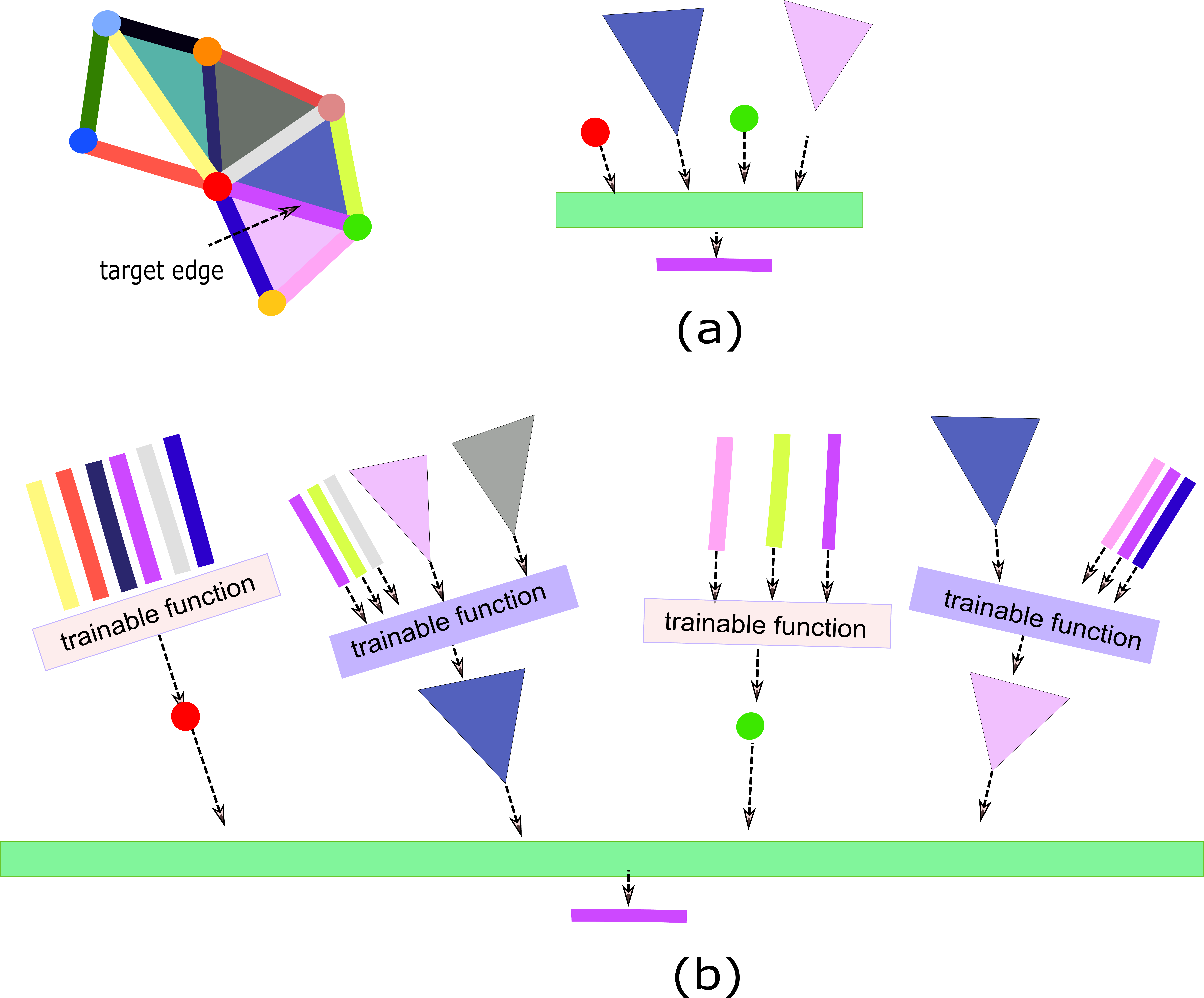}
    \caption{Homology and cohomology message passing scheme. The figure illustrates the computation of a two layers cell complex network with respect the specified edge.}
  \label{DMF3}}
\end{figure}

Finally, the \textit{Homology and Cohomolgy Message Passing Scheme} (HCMPS) is given by
\begin{equation}
\label{hcmps}
    H^{(k)}_{m}:=M\left( \partial,\mathbf{\partial}^T, H^{(k-1)}_{m-1},H^{(k-1)}_{m+1} ;\theta^{(k)}_{m}\right) ,
\end{equation}
where $0 \leq m \leq n-1$, $1 \leq k \leq L$,  $H^{(k)}_{m} \in \mathbb{R}^{|X^{m}| \times d_k}  $ are the embeddings computed after $k$ steps of applying Equation \eqref{hcmps}, $\partial,\mathbf{\partial}^T$ are the boundary and coboundary operators of the input complex.  An example of applying the HCMPS is given in Figure \ref{DMF3}.

\section{Entire Simplicial Complex Learning  }
\label{SCA}
Our proposed method for learning entire simplicial complex representation relies on collecting node-level simplicial representation and combining them together in order to obtain simplicial complex-level representation. In this section, we review the AMPS-simplicial autoencoder introduced in \cite{hajijcell}. We then show how it can be utilized to obtain an entire-level simplicial representation by using metric learning. 

\subsection{Simplicial Complex AutoEncoders (SCAs)}

\label{autoencoders}

Let $X$ be a simplicial complex of dimension $n$. \textit{A simplicial complex autoencoder} (SCA) on $X$ consists of three components,
namely of an encoder-decoder system, a user-defined similarity measure, and a user-defined loss function
\cite{hamilton2017representation,hajijcell}.

Firstly, the SCA encoder-decoder system is described. The encoder is a function of the form :
$enc:X^{<n}\to \mathbb{R}^d$ and it associates to every $k-$ simplex $a^k$ in $X$ an embedding $\textbf{z}_{{a}^k}$ in $\mathbb{R}^d$.
The decoder is a function of the form :
$dec : \mathbb{R}^d\times \mathbb{R}^d \to \mathbb{R}^+$ and it associates to every pair of simplex embeddings $( \textbf{z}_{a^k}, \textbf{z}_{c^l})$ a measure of similarity $dec( \textbf{z}_{a^k} \textbf{z}_{c^l})$ that quantifies some notion of relationship between $a^k$ and ${c^l}$. The functions $enc$ and $dec$ are trainable functions. In particular, the encoder can be chosen to be a cell complex network as illustrated in Section \ref{cxn}.

Secondly, a user-defined similarity measure is required by a SCA. We seek to train the encoder-decoder functions such that the trained similarity is as close as possible to the user-defined similarity: $dec( enc(a^{k}), enc(c^l)) = dec( \textbf{z}_{a^k}, \textbf{z}_{c^l})\approx sim_{X}(a^{k},c^l),$ where  $sim_X: X^{<n}\times X^{<n} \to \mathbb{R}^+$ is a user-defined function such that $sim_X(a^k,c^{l})$ reflects a user-defined similarity between the two simplicies $a^{k}$ and $c^{l}$ in $X^{<n}$. For instance, the similarity measure function on $X$ can be simply chosen the adjacency matrix $A_{adj}$ defined in Section \ref{cxn}.

Thirdly, a user-defined loss function is required by a SCA. Training the encoder-decoder system is done by specifying a loss function $l : \mathbb{R} \times \mathbb{R}\to \mathbb{R}$ and by defining
\begin{equation}
\label{loss}
    \mathcal{L}_k := \sum_{\mathcal{CO}[a^k,c^k] }
    l\left(  dec(  enc(\textbf{z}_{a^{k}}), enc(\textbf{z}_{c^k})),sim(a^{k},c^k)\right)
    \end{equation}
and $\mathcal{L}:=\sum_{k=0}^{n-1} \mathcal{L}_k$.
The sum in Equation \eqref{loss} is taken over all possible $\mathcal{CO}[a^k,c^k]\subset X^{k+1}$.

Table \ref{tab:title} shows several concert methods to define the autoencoders on simplicial complexes. Observe that any variant of the geometric message passing protocols can be used with any of the proposed simplicial complex autoencoders.

.\begin{table}[h] \centering \caption{Various definitions of simplicial somplex autoEncoders. } \label{tab:title} \begin{adjustbox}{width=\columnwidth,center} \begin{tabular}{||c|c|c|c||} \hline Method &  Decoder & similarity & Loss \\ [0.5ex] \hline Laplacian eigenmaps \cite{belkin2001laplacian}  &  $||\textbf{z}_{a}-\textbf{z}_{c}||_2^2$ & general & $dec(\textbf{z}_{a},\textbf{z}_{c}).sim(a,c) $ \\ Inner product methods \cite{ahmed2013distributed} & $ \textbf{z}_{a}^T\textbf{z}_{c}$ & $A_{adj}(a,c)$ & $||  dec(\textbf{z}_{a},\textbf{z}_{c})-sim(a,c) ||_2^2$ \\ Random walk methods \cite{grover2016node2vec,perozzi2014deepwalk} & $  \frac{ e^{\textbf{z}_a^T \textbf{z}_c}}{\sum_{b\in X^{k}  } e^{\textbf{z}_a^T \textbf{z}_b } } $ & $p_X(a|c)$   & $  -log(dec(\textbf{z}_{a},\textbf{z}_{c})) $ \\ [1ex] \hline \end{tabular} \end{adjustbox} \end{table}

\subsection{Learning Entire Simplicial Complex Embedding}
\label{learning entire embedding}

Let $enc:X^{<n}\to \mathbb{R}^d$ be a simplicial complex encoder. Denote by $ \mathcal{U}_X \in\mathbb{R}^{\hat{N}\times d}$ to the simplices embeddings of $X^{<n}$ that are induced by the function $enc$. Our proposed method relies on learning a weighted sum of the simplex-level representations encoded in $ \mathcal{U}_X$. Specifically, we seek to learn a simplicial complex-level embedding of the form
\begin{equation}
\label{eq:h}
   \mathbf{h}_{X}=\sum_{m=1}^{\hat{N}} w_{m}(\mathcal{U}_X;W) \textbf{z}_m ,
\end{equation}
where $w_{m}(\mathcal{U}_X;W) \in \mathbb{R} $ is a weight of the simplex embedding $\textbf{z}_m$ that depends on $\mathcal{U}_X$ and parametrized by $W \in \mathbb{R}^{d\times d } $, a trainable weight matrix \footnote{Note that in Equation \eqref{eq:h}, we did not include the dimension of simplex embedding in the training. This restriction is not needed and we are only making this assumption for notational convenience.}. The weight $w_{m}$ can be chosen in many different ways, here we simply follow \cite{bai2019unsupervised} and define the weight as
\begin{equation}
   w_{m}(\mathcal{U}_X;W)= \sigma\left((\textbf{z}_m)^T \relu\left(W \sum_{n=1}^{\hat{N}} \textbf{z}_n\right) \right), 
\end{equation}
where $\sigma(x)=\frac{1}{1+exp(-x)}$. Finally, the embedding $\mathbf{h}_X$ can be learned in multiple ways. For instance, given a collection of simplicial complex $\{X_i\}_{i=1}^m$ one may learn complex-to-complex proximity embeddings by minimizing the objective
\begin{equation}
    \mathcal{L}= \sum_{i=1}^m \sum_{j=1}^m 
    \left(||h_{X_i} - h_{X_j}|| - d_{ij} \right)^2,
\end{equation}
where $D=[d_{ij}]$ is an appropriately chosen distance matrix on the simplicial complexes  $\{X_i\}_{i=1}^m$. For example, the Haussdorf distance on simplicial complexes \cite{marin2020hausdorff} can be employed to compute the distance matrix $D$. Alternatively, in special case when the simplicial complex is a triangulated mesh, more efficient methods to compute the metrics can be utilized such as persistence homology-based metrics \cite{hajij2020fast,zhang2019mesh} or Laplacian-based methods \cite{crane2013geodesics}.

\section{Experiments}
\label{test}
\begin{figure*}[ht]
  \centering
   {\includegraphics[scale=0.35]{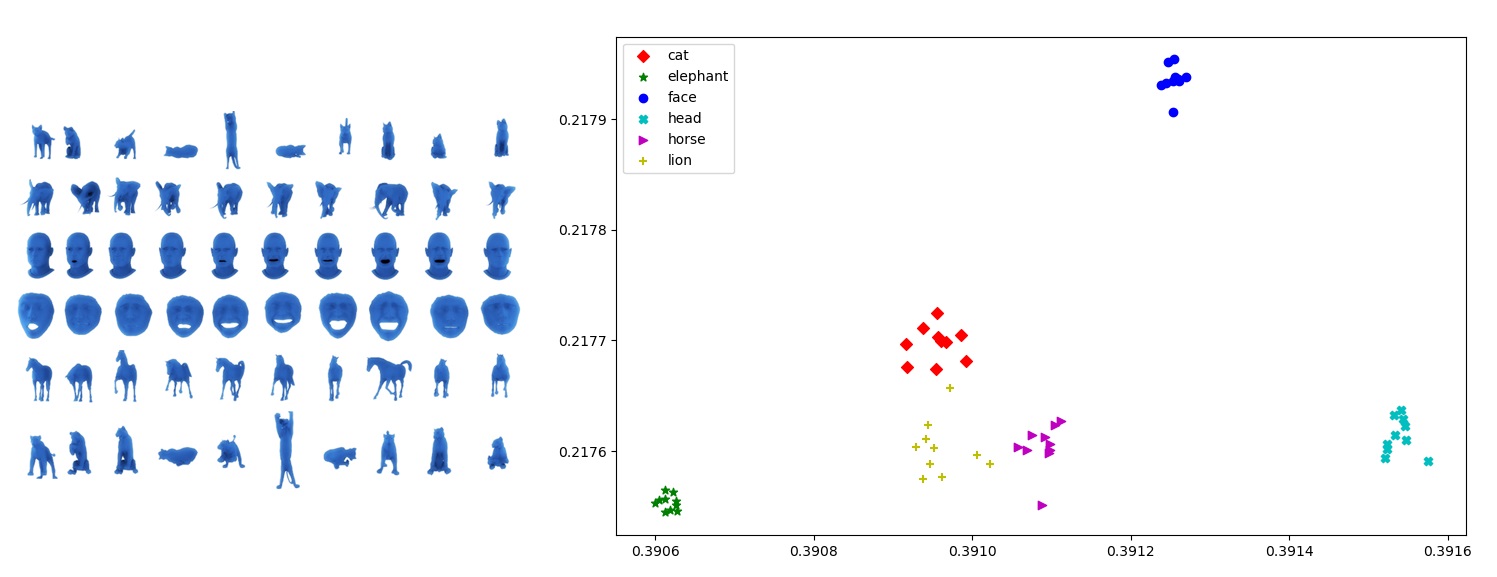}
    \caption{Initial results using our proposed simplicial complex autoencoder. The figure shows the entire simplicial complex embeddings of the whole dataset. The final embedding was chosen to be in $\mathbb{R}^2$. The model was trained on faces, heads, and horses. The inference indicated in the figure shows the entire embeddings for the 60 meshes. Note that while the model was never trained on the cat, camel and elephant meshes, it succeeded in clustering these meshes. Moreover, all animals with similar shapes are clustered together.  }
  \label{final_results}}
\end{figure*}

To test our proposed method, we train an AMPS-SCA on a publicly available datset. Initially, we experiment on a mesh dataset to help making a visual inspection of the SCA performance. This dataset, which is described in details in \cite{sumner2004deformation}, consists of 60 meshes that belong to 6 categories: cat, elephant, face, head, horse, and lion. 
Each category contains ten triangulated meshes. We train the model on the faces, heads, horses (total of thirty meshes); the rest of the dataset (i.e., elephants, horses and cats) were used to for testing. The model was trained using our PyTorch-based implementation. The final model consist of an AMPS single layer model that we described in Equation \eqref{CCXN} with one additional dense layer at the end that operates on the vector $\mathbf{h}_X$. The model was trained using SGD techniques for $100$ epochs with a batch size of $1$; roughly, the model sees each model in the training dataset around three times.
The final embedding dimension of the input mesh was chosen to be $2$ for visualization purposes. The results are shown in Figure \ref{final_results}. From the figure, we can observe that the model successfully clusters meshes with the same category close to each other. Moreover, meshes with related structure, quadruple animals, are also embedded in a spatially close neighborhood.

\section{Discussion}
In Section \ref{learning entire embedding}, we describe one way to learn the embeddings $\mathbf{h}_X$ in an unsupervised fashion. The method learns the proximity between simplical complexes that is encoded in a pre-computed matrix $D$ on a dataset of simplicial complexes. The problem with this method is that it requires the computation of the entire distance matrix which might be computationally inefficient. There are many other potentially good methods to learn such embeddings in an end-to-end fashion. For instance, a potential method to learn the metric of simplicial complex embeddings can be done by utilizing triplet loss method as proposed in \cite{hoffer2015deep}. Metric learning with the triplet loss method construct a triplet net which consists of three shared parameter feedforward networks. The network is fed three embeddings $\mathbf{h}_{X}$, $\mathbf{h}_{X^+}$ and $\mathbf{h}_{X^-}$ where $\mathbf{h}_{X}$ and $\mathbf{h}_{X^+}$ are of the same class, and $\mathbf{h}_{X}$ and $\mathbf{h}_{X^-}$ are of different class. We stress here the fact that while a binary label is utilized when choosing the triplet $(X,X^+,X^-)$, triplet network can effectively learn a metric and determine which simplicial complexes are closer to a given complex $X$. In other words, the interpretation of sharing the same class is correlated with embedding closeness in the embedding space \cite{hoffer2015deep}. The advantages of the triplet loss method is that we do not need to pre-compute a distance matrix on the simplicial complexes training set. However, the triplet loss method requires a labeled training dataset which might not be always available.

\section*{Acknowledgment}
M.H. was supported in part by the National Science Foundation (NSF, DMS-2134231). Although the current affiliation of G.Z. is the National Institutes of Health (NIH), this work was designed and implemented while G.Z. being with the University of South Florida. The opinions expressed in this article are the author's own and do not reflect the view of NIH, the Department of Health and Human Services, or the United States government. VM would like to acknowledge partial support for this project by the ARO contract number W911NF-21-1-0094.


\bibliography{iclr2021_workshop}
\bibliographystyle{iclr2021_workshop}


\end{document}